\documentclass[letterpaper, 10 pt, journal, twoside]{IEEEtran}
\usepackage{times}
\usepackage[normalem]{ulem}
\usepackage{subfiles}
\usepackage[numbers]{natbib}
\usepackage{multicol, multirow}
\usepackage[bookmarks=true]{hyperref}
\usepackage{balance}
\usepackage{siunitx}
\usepackage{xspace}
\usepackage{tabularx, graphicx, amsmath, amsfonts, pifont, amssymb, subcaption}
\usepackage{booktabs}
\usepackage{pdfpages}

\usepackage[some]{background}

\SetBgScale{0.8}
\SetBgContents{\parbox{20cm}{%
   \centering \Large This paper has been accepted for publication at the \\ 
   Robotics: Science and Systems (RSS), Daegu,  Republic of Korea, 2023\\[32cm]}}
\SetBgColor{gray}
\SetBgAngle{0}
\SetBgOpacity{0.7}

\newcommand{\sota}{state-of-the-art\xspace}

\begin{document}

\title{InstaLoc: One-shot Global Lidar Localisation in Indoor Environments through Instance Learning}

\author{Lintong Zhang$^{1}$, Tejaswi Digumarti$^{1}$, Georgi Tinchev$^{2}$, and Maurice Fallon$^{1}$\\
 $^1$\,University of Oxford \hspace{12pt} $^2$\,Amazon Research\\[2pt]
}

\BgThispage

\maketitle

\begin{abstract}
Localization for autonomous robots in prior maps is crucial for their functionality. This paper offers a solution to this problem for indoor environments called InstaLoc, which operates on an individual lidar scan to localize it within a prior map.
We draw on inspiration from how humans navigate and position themselves by recognizing the layout of distinctive objects and structures. Mimicking the human approach, InstaLoc identifies and matches object instances in the scene with those from a prior map. As far as we know, this is the first method to use panoptic segmentation directly inferring on 3D lidar scans for indoor localization. 
InstaLoc operates through two networks based on spatially sparse tensors to directly infer dense 3D lidar point clouds. The first network is a panoptic segmentation network that produces object instances and their semantic classes. The second smaller network produces a descriptor for each object instance. A consensus based matching algorithm then matches the instances to the prior map and estimates a six degrees of freedom (DoF) pose for the input cloud in the prior map. The significance of InstaLoc is that it has two efficient networks. It requires only one to two hours of training on a mobile GPU and runs in real-time at \SI{1}{\hertz}. Our method achieves between two and four times more detections when localizing, as compared to baseline methods, and achieves higher precision on these detections.
\end{abstract}

\IEEEpeerreviewmaketitle

\section{Introduction}

Localization is a fundamental capability needed for mobile robots to navigate their environment and make decisions. There have been many studies on vision, lidar, and radar-based localization. The parent problem of Simultaneous Localisation and Mapping (SLAM) concerns a robot determining its pose while building a map of its environment concurrently. Localization, or place recognition, can contribute to SLAM by helping to \textit{close loops}, or to determine the robot's position in a fixed prior map - the \textit{kidnapped robot} problem.

Many popular localization methods using visual and lidar sensors have been proposed. Among visual-based approaches, visual teach-and-repeat \cite{mactavish2018selective, matias2021visualTandR} is one of the most popular methods, where a robot first constructs a visual prior map and then localizes on its repeat phase. Compared to image-based camera solutions, modern 3D lidar sensors are view-invariant, robust to lighting changes, and can operate when the path traveled is offset from the original path. Given that lidar is a precise and long-range sensor, lidar localization has been heavily researched in outdoor environments, especially in the context of autonomous driving \cite{Vidanapathirana2020LocusLP, Kong2020SemanticGB, zhu2020gosmatch}. However, there are fewer approaches for indoor environments because these environments contain more complex structures and clutter, hence fewer clear separations between objects in lidar scans. In an indoor environment, there are many different classes of objects, with one dataset proposing 13 semantic classes ~\cite{s3dis2016}. The indoor scene varies greatly: from bare box-shaped rooms with four walls to narrow corridors with two long walls. Room surfaces are often covered with objects such as electronics, hanging art, ceiling lights, bookcases, and various decorative objects. Localization algorithms cannot rely on flat ground assumptions as there is often an incomplete view of the floor. In addition, there are changes in levels, with steps and staircases. Nevertheless, it is important to localize in these indoor scenarios to enable robots to operate robustly in complex office buildings, construction sites, warehouses, and other commercial environments.

\begin{figure}[!t]
    \centering
    \includegraphics[width=\columnwidth]{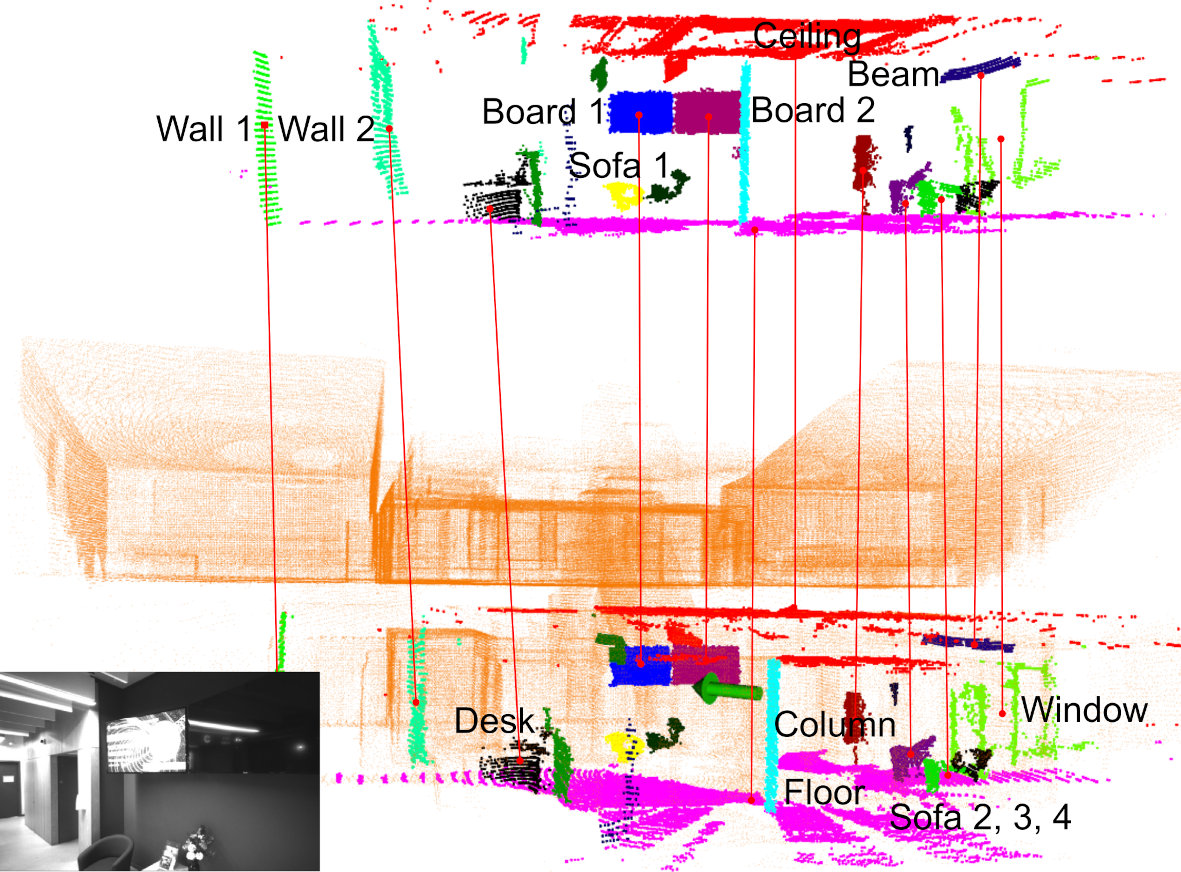}
    \caption{An illustration of the \textit{InstaLoc} method, where a `live' lidar scan (top) is localized inside the orange colored prior map (bottom) by matching semantically segmented objects (red lines). The green arrow shows the estimated position and the left corner image is the corresponding camera view.}
    \label{fig:teaser}
\end{figure}

In this paper, we draw inspiration from how humans perceive the world and reach the ``I know where I am'' moment. By memorizing and recognizing the distinctive structures and unique objects inside a space, humans can spatially locate themselves in the environment. Based on the same principle, InstaLoc makes use of individual object instances to localize. Different from existing approaches that rely on primitive shapes or other handcrafted features, InstaLoc learns to segment and match individual objects to the prior scene. These scenes include both fixed objects (walls, ceilings, beams) and movable objects (chairs, desks), in order to tolerate active dynamics and longer term scene changes.
To train the segmentation network with accurate class labels, we leverage a simulator to synthesize and automatically annotate every point --- thus avoiding onerous point cloud labeling. To overcome the challenge of imperfect instance segmentation, we designed a sparse convolutional descriptor network that infers many instances simultaneously and tolerates mild changes in the instance point cloud.

To summarize, our contributions are:
\begin{itemize}
    \item A novel learning-based lidar localization approach for indoor environments that can process dense lidar scans on a mobile GPU in real time.
    \item An improved panoptic segmentation network that works with single lidar scans.
    \item A fast and efficient descriptor network to learn object instances with a variable number of input points.
    \item State-of-the-art performance on indoor localization compared to other segment-based methods, achieving two to four times more detection. 
\end{itemize}

\section{Related Work}

In this section, we describe recent work on segment-based lidar localization and its applications to urban, natural, and industrial environments. We review approaches that use semantic segmentation for outdoor localization. Finally, we discuss methods that rely on the geometry of the scene and algorithms which can localize in maps made with other sensor modalities (co-localization).

\subsection{Outdoor Segment-based localization}
Scan segment matching was first introduced by ~\citet{douillard2012scan}, where segments were considered as a midway point between local and global approaches for describing a scene. The approach was initially applied to lidar localization by ~\citet{segmatch2017} where segments were extracted directly from raw point clouds and described with a descriptor based on the geometry of the segment (such as its eigenvalues and proportions). Later, ~\citet{Dub2018SegMapRSS,Tinchev2019esm} described segments using a neural network to provide a richer and more meaningful descriptions. Building on this research, ~\citet{Ratz2020OneShotGL} showed that lidar segments fused with visual data further improve the performance of global localization algorithms. ~\citet{Cramariuc2021SemSegMap3} fused both colour and semantic information from images to create an enriched point cloud that was later segmented and used for localization. We use this segment-direct concept as the basis of InstaLoc, however in contrast to these approaches, InstaLoc does not use engineered segmentation methods, nor images, to extract the semantic information but directly learns to predict the per-point instance annotation.

There are several relevant outdoor lidar localization methods that make use of semantics and segments. ~\citet{Vidanapathirana2020LocusLP} used global descriptors with segments and spatiotemporal high-order pooling for place recognition. ~\citet{Kong2020SemanticGB} presented a semantic graph-based approach to place recognition, where the topological information of the point cloud is preserved. ~\citet{zhu2020gosmatch} extracted common semantic classes, such as vehicles, trunks, and poles from the raw point cloud for loop closure detection. The above methods are designed primarily for outdoor scenarios and are inadequate for an indoor setting. However, they demonstrate the value that semantic information brings to place recognition. To extend this line of research, we leverage a panoptic segmentation method that predicts both the semantic mask and instance label of each point.

\subsection{Indoor localization}
Specifically focusing on indoor localization, the \sota methods often focus on planar floors or geometric features which describe corners and intersections as landmarks for localization. For example,~\citet{GCLO} used planar floor assumption to constrain the vertical pose drift of a robot in a multi-floor parking lot. ~\cite{piLSAM,LIPS} used planar surfaces to efficiently align two lidar scans for loop closure detection. ~\citet{floorSGD,GLFP} used floor plan features such as corners and wall intersections for localization. ~\citet{2DlidarSemanticIndoor} proposed to use semantic features to detect and match corners of doors and walls. Other works rely upon a predefined map of the world such as a BIM model or a floor plan.~\citet{SemanticBIM2D, semanticBIM3D} built a map from a subset of semantic entities and their associated geometries drawn from a BIM model of the world. They used a spatial database to query the position of the robot within a graph-based localization approach. They impose a prior to use static features for localization. In comparison to these approaches, we do not rely on planes or any other explicit structure to constrain our localization performance. Instead, our approach is to learn to segment semantically meaningful objects and match them between different observations of the scene.

\section{Methodology}

\begin{figure*}
    \centering
    \includegraphics[width=\linewidth]{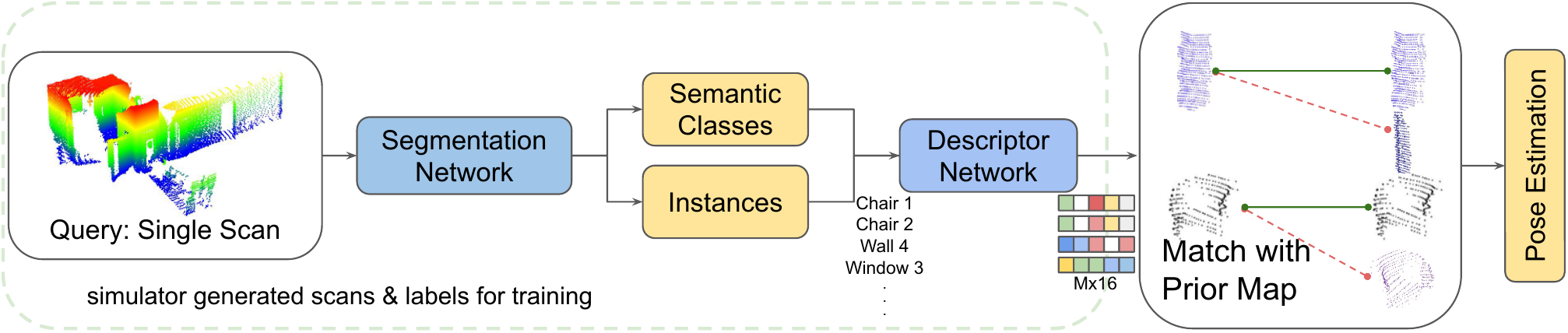}
    \caption{Overview of the proposed learned lidar localization system}
    \label{fig:overview_detailed}
\end{figure*}

In this section, we first formulate the research problem, then present the entire pipeline as shown in Fig.\ref{fig:overview_detailed}: the panoptic segmentation network, the instance description module, and the matching and pose estimation module, 

\subsection{Overview}
\label{sec: overview}
The problem is defined as localizing a single query lidar scan $\mathcal{Q} = \{\boldsymbol{q}_i \in \mathbb{R}^3 \}$ within a prior map $\mathcal{M} = \{P_1, P_2, \dots P_{t_{i-1}}\}$.
\noindent We seek to determine the pose of the lidar at time $t_i$ defined as follows,
\begin{equation}
\boldsymbol{x}_i \triangleq \left[ \mathbf{t}_i, \mathbf{R}_i \right] \in \ensuremath{\mathrm{SO}(3)}\xspace \times \mathbb{R}^{3}
\end{equation}
where $\mathbf{t}_i \in \mathbb{R}^3$ is the translation, $\mathbf{R}_i \in \mathrm{SO(3)}$ is the orientation of $\mathcal{Q}$ in $\mathcal{M}$. The map $\mathcal{M}$ is a collection of registered lidar scans, $P_t = \{\boldsymbol{m}_{i,t} \in \mathbb{R}^3 \}$, accumulated over time.

We approach the problem at the level of objects and compute the pose $\boldsymbol{x}_i$ by matching object instances identified in the query scan $\mathcal{Q}$ with those previously identified in the map $\mathcal{M}$.

The first step is to partition the map scans into meaningful object instances.
Prior approaches have used planes or region growing methods to segment objects with a scan.
This segmentation approach works well in outdoor environments such as in the case of autonomous vehicle localization.
This is because sizes of outdoor objects and the separation between them is many times greater than the average inter-point distance in a lidar scan.
However, in indoor environments, due to the close proximity of objects with each other, space partitioning and region growing approaches perform poorly.
On the other hand, there are many distinguishable objects such as furniture, doors and windows; because of this we can use  semantic object segmentation to partition the environment into these object segments.

Objects observed in a query scan will often be quite different from those in the prior map.
This could be due to observing the object from a different viewpoint in the query scan than from which it was observed in the prior map. Partial observations from different viewpoints, occlusions by other objects, or different point sampling densities ( if the query and map scans were taken from different ranges) all contribute to variation in the reconstruction of an object instance.
Due to this variation in the observations, finding matches by aligning a 3D point cloud of the objects between the query and the map will result in poor pose estimation.
To overcome this issue, we use object level descriptors that capture the distinguishing features of each object.
Estimating pose by matching these descriptions provides some robustness against the variations which occur due to differing viewpoints and partial observations.
Descriptor matching also requires lower computational and memory resources as the dimensions of the descriptor are typically smaller than the number of points in each object.

After this step, descriptors of the objects segmented from the query scan need to be matched against the database of objects with descriptors in the map to determine correspondences between the query scan and the map.
We use the approach from ~\cite{Aldoma2012AGH} to group descriptors based on their similarity and to find correspondences.
Finally, we use RANSAC on a subset of correspondences to estimate the 6-DOF pose of the lidar sensor by aligning the matched objects between the two scans.

In InstaLoc both the instance segmentation module and the instance description module are modeled using deep neural networks which work directly on 3D point cloud data.
Typical lidar scans are point clouds with large amounts of spatially sparse data.
We use sparse tensors to represent this data and designed both networks in our framework using the Spatially Sparse Convolution Library (SpConv) \cite{spconv2022} which uses sub-manifold sparse convolutions in its neural network implementation.
Sub-manifold convolutions have the advantage that they maintain a greater degree of sparsity than other sparse convolutions by overcoming the issue of sub-manifold dilation~\cite{graham20183d}.
As a result, deeper networks with lower memory and computational requirements, and practical real-time capabilities can be constructed to work with large amounts of sparse data.
Furthermore, ~\citet{graham20183d} also showed that sub-manifold sparse convolutions are more efficient than alternate approaches that use spacial partitioning schemes.

\subsection{Instance Segmentation}
The instance segmentation module is a point-wise panoptic segmentation network.
Given a lidar scan, i.e. a set of $N$ 3D points, $P = \left\{\boldsymbol{p}_1, \boldsymbol{p}_2, \dots \boldsymbol{p}_N | \boldsymbol{p}_k \in \mathbb{R}^3\right\}$ as input, the network predicts for each point $\boldsymbol{p}_k$ a semantic label $s_k$ corresponding to the object class that the point belongs to (e.g. chair, table, wall, ceiling) and an instance label $i_k$ representing the unique object that the point corresponds to (e.g. chair1, chair14 or chair42).
We use the state-of-the-art \textit{Softgroup} \cite{Vu2022SoftGroup} network architecture to construct this module. 
This architecture consists of three stages; (1) a U-Net based point-wise prediction network that generates semantic scores and an offset vector representing the distance from the point to the instance it belongs to, (2) a soft-grouping step where points are grouped by similarity of their semantic scores and their spatial distance to generate instance proposals and (3) a refinement network that extracts features for every instance proposal and then uses a tiny U-Net based network to refine the proposals.

A fixed distance threshold used to group the points in step (2) of the \textit{Softgroup} architecture works well for point clouds with uniform sampling density.
In lidar scans, the sampling density is much lower along the vertical axis as compared to the density in the horizontal axis and points become further separated at the sensing range increases.
If a fixed distance threshold is used for grouping, then the number of instance proposals would be overestimated in regions further away from the sensor; this can lead to incorrect object segmentation.
We counter this issue by using an adaptive radius threshold proportionate to the vertical distance between two beams.
Typically, 3D lidars rotate \SI{360}{\degree} horizontally and have a vertical field of view of $\theta$ radians. 
For a point $\boldsymbol{p}_i(x_i, y_i, z_i)$ resulting from a lidar beam in a point cloud, with the sensor origin $O$, its radius threshold $\rho_i$ is:
\begin{equation}\label{eq: adaptive-theshold}
    \rho_i = \alpha \cdot d(\boldsymbol{p}_i, O) \cdot \tan(\frac{\theta}{N_{beam}})
    \\
    =\alpha \cdot d(\boldsymbol{p}_i, O) \cdot \frac{\theta}{N_{beam}}
\end{equation}
where
\begin{equation}
    d(\boldsymbol{p}_i, O) = \sqrt{x_i^2 + y_i^2 + z_i^2}
\end{equation}
and $N_{beam}$ is the number of lidar beams and $\alpha$ is a constant scale factor. 

The output of panoptic segmentation network is a set of $M$ object instances $\mathcal{I} = \{ I_1, I_2, \dots, I_M \}$ where each object instance is a set of $N_j$ points representing the 3D coordinates of the point and the semantic label $s_j$ of the object, i.e. $I_j = \left\{ \boldsymbol{h}_{k,j} | k=\{1, 2 \dots N_j\}, \boldsymbol{h}_{k,j} = (\boldsymbol{p}_{k,j}, s_j) \right\}$.

\subsection{Instance Description}
\begin{figure}
    \centering
    \includegraphics[width=\columnwidth]{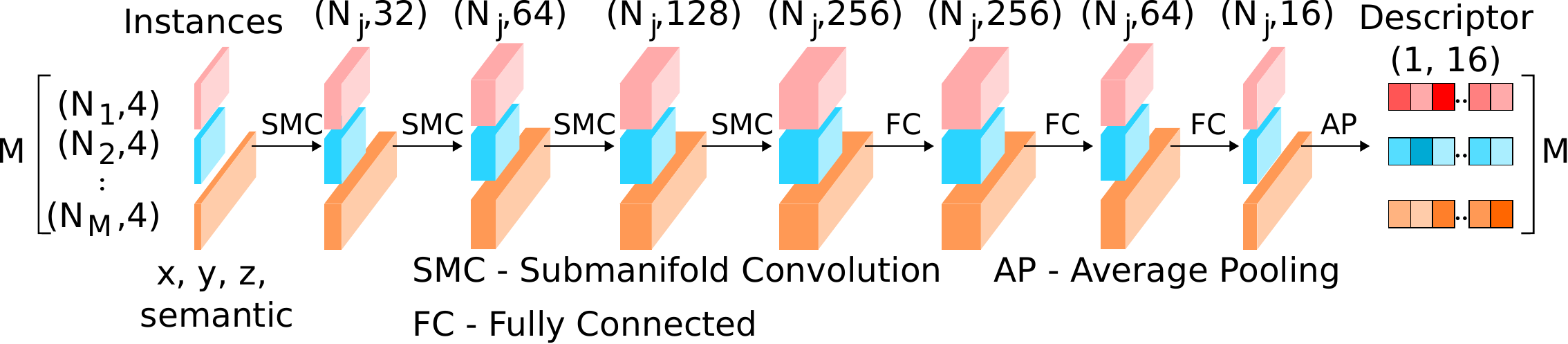}
    \caption{Instance descriptor network architecture. The input is a set of object instances with a variable number of points per instance, with each point representing the 3D coordinates of the point and the semantic label of the object. The network consists of layers of sub-manifold convolutions with increasing feature size followed by fully connected layers, with dropout before the final fully connected layer. Finally an average pooling layer computes a single descriptor for each object instance.}
    \label{fig:descriptor-net}
\end{figure}

After the object instances are segmented, the next step is to generate descriptors for each of the instances. 
An overview of the network is shown in Fig. \ref{fig:descriptor-net}. 
The network is designed to be small and fast: it can take all instances in one batch with varying number of points as input.
This is done using the instance descriptor network, which consists of four sub-manifold sparse convolutional layers of increasing feature size followed by three fully connected layers, with a dropout layer before the final fully connected layer.
The input to the network is a set of object instances $\mathcal{I} = \{ I_1, I_2, \dots, I_M \}$, the output of the instance segmentation network, with each instance $I_j$ containing $N_j$ points (with $N_j$ varying for each object).
The descriptor network output for each object instance $I_j$ is an $N_j \times D$ tensor where every row is a descriptor of length $D$ for one point in the object instance.
Finally, an average pooling layer computes the average of the $N_j$ descriptors to create a single descriptor of length $D$ for each object instance.
This results in an output vector of dimensions $M \times D$, where $M$ is the number of object instances.

The network is trained using triplet loss. If $a, p, n \in \mathbb{R}^D$ are the descriptors for an anchor, the corresponding positive element and a negative element respectively, then the triplet loss $\mathcal{L}_{triplet}$ can be calculated as
\begin{equation}\label{eq: triplet-loss}
    \mathcal{L}_{triplet}(a, p, n) = \max\left\{d(a, p) - d(a, n) + m, 0\right\}
\end{equation}
where
\begin{equation*}
    d(x, y) = ||x - y||_2
\end{equation*}
is the pairwise distance between the descriptors; and the margin $m$ is set to 1. 
The average loss over all the samples in a mini-batch is considered as the loss during training.

\subsection{Matching}
For each instance in the query scan, we first obtain its $N$ closest descriptors from the database of instances of the prior map. 
This generates a list of instance-to-instance correspondences. 
A correspondence grouping method from \cite{Aldoma2012AGH} is used to find the correct correspondence. 
We start from a seed correspondence $c_n = \{I_n^Q, I_n^M\}$, where $I_n^Q$ and $I_n^M$ are two instances from the query scan and the prior map respectively. 
We then loop through all candidate correspondences, so another correspondence $c_m = \{I_m^Q, I_m^M\}$ can be grouped with $c_n$ if: 
\begin{equation}
||I_n^Q - I_m^Q|| - ||I_n^M - I_m^M|| < \epsilon
\end{equation}
$\epsilon$ is the parameter that restricts how strictly the grouping algorithm behaves. The accepted consensus group has to contain a minimum of $\tau$ instances. Finally, for the 6 DoF pose estimation, we apply a RANSAC step on the subset of correspondences to align the query scan with the prior map, with $\tau$ and $\epsilon$.

\section{Implementation}

\subsection{Simulated Lidar Data}

\begin{figure}
\centering
\includegraphics[width=0.48\columnwidth, height=3cm]{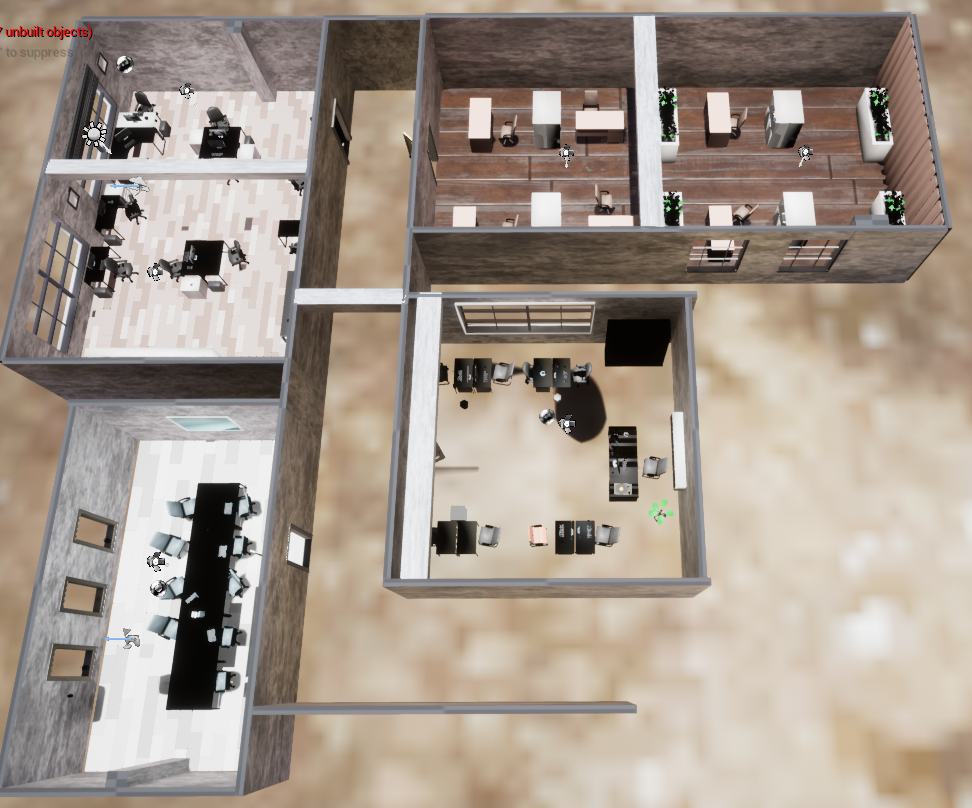}%
\hspace{1mm}%
\includegraphics[width=0.48\columnwidth, height=3cm]{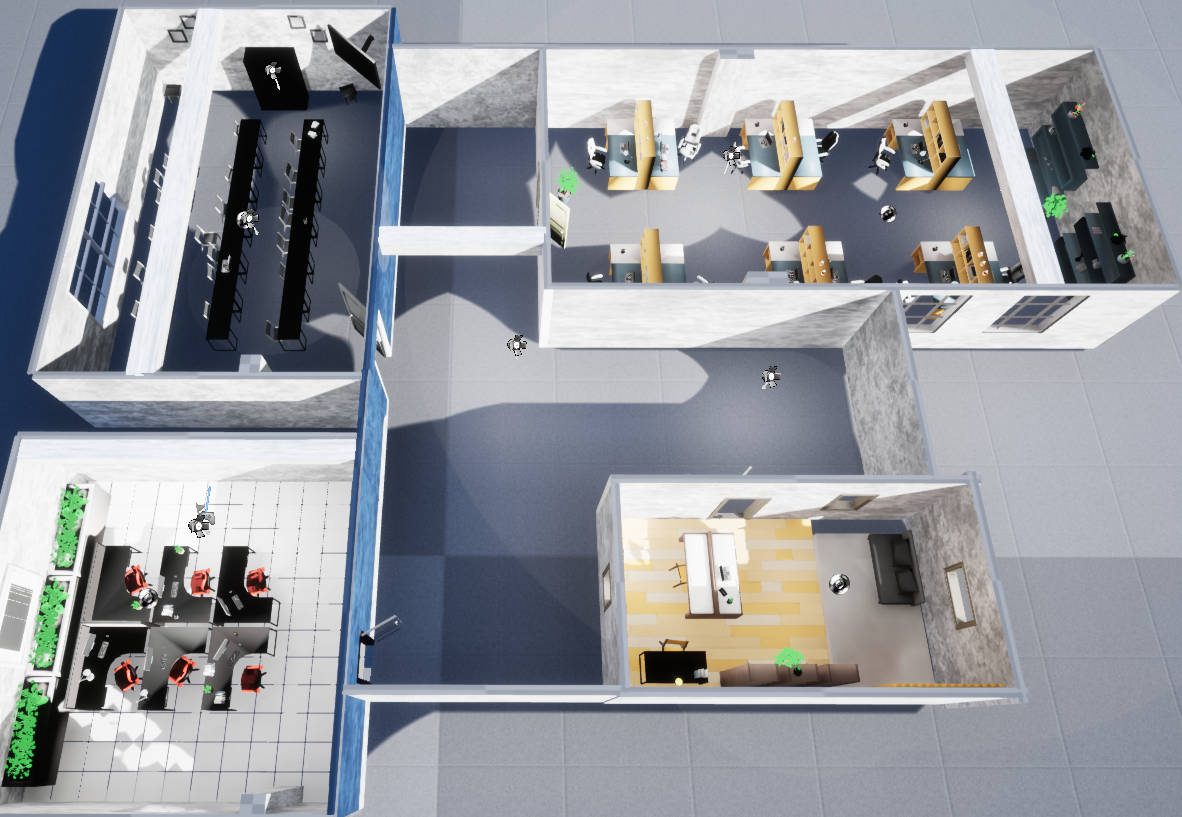}\\
\caption{Two indoor office networks constructed using the Unreal Engine to simulate lidar scans with semantic labels.}
\label{fig:UE-spaces}
\vspace{-1mm}
\end{figure}

Training deep learning algorithms requires large amounts of data.
To bypass the need to do time-consuming manual labeling, we constructed several indoor environments in the Unreal Engine game simulator to take advantage of automatic labeling. As well as being automatic, it eliminates errors in human labeling and can be easily extended to other environments. We created about 20 unique rooms and assembled them into six room networks which contained a total of $\sim$1500 objects. As an example, two of the six networks are shown in Fig. \ref{fig:UE-spaces}. We used the Airsim plugin \cite{Shah2017AirSimHV} to capture over 90 scans from these spaces.

The simulator allowed us to configure the lidar settings --- including frequency, range, the field of view, and the number of lidar beams. The simulated lidar configuration we used was modelled on the Ouster OS-128 lidar \footnote{\url{https://ouster.com/products/scanning-lidar/os0-sensor/}}, which has $\sim$\SI{50}{\meter} range, \SI{90}{\degree} field of view, and 128 lidar beams. Note, that this is a wide field of view and dense lidar coming on the market. Similarly to the existing indoor point cloud dataset, Stanford 3D Indoor Scene Dataset (S3DIS) \cite{s3dis2016}, we used 13 object classes: ceiling, floor, column, beam, wall, table, chair, bookcase, sofa, window, door, board, and clutter.

\subsubsection{Labeled Data for Instance Segmentation}
 Each simulated lidar beam that intersects with an object would result in a range measurement and a unique object ID. Using the object ID, we can assign a semantic class and an instance number. These labels are used in the supervised training of the two networks. Overall, each point has five fields: $(X, Y, Z)$ coordinates, semantic class, and object instance number. 

\subsubsection{Triplets for Descriptor Network}
To train the descriptor network, we need to generate object instances as triplets - with anchor, positive and negative instances. First, we generate two scans that are 2m apart and a \SI{10}{\degree} rotation in the simulator. Given that every object in the lidar scan is labeled, we classify the same objects in these two scans as the anchor and positive instance. We then randomly selected another object as the negative instance. Because the anchor and positive instances have mild viewpoint differences, the objects scanned in the point clouds may have slight changes. These slight appearance changes contribute to algorithm robustness. In total, about 9900 triplet object instances were generated for training, validation, and testing. 

\subsection{Training}
As mentioned in Sec. \ref{sec: overview}, both networks are built with a sparse tensor framework and were trained on a 4GB mobile GPU, NVIDIA Quadro T2000.
\subsubsection{Instance Segmentation Network}
We use a pre-existing Softgroup model (trained on S3DIS) as a warm start. The voxel size was set to \SI{2}{\cm} and the minimum number of points in each instance was set to 50. The network was trained for 50 epochs which took about one hour.
\subsubsection{Instance Descriptor Network}
The network is trained from scratch with a triplet loss function, see \eqref{eq: triplet-loss}. Compared to a whole scan (which usually contains over 100,000 points) each triplet instance is only a small fraction of a whole scan; because of this we could increase the batch size to allow parallel input. The descriptor network was trained for 90 epochs, and took around 90 minutes.

Both networks were trained with an Adam optimizer with a learning rate of 0.001.

\section{Experiment and Results}

\begin{figure*}
\centering
\vspace{1mm}
\includegraphics[width=0.85\linewidth]{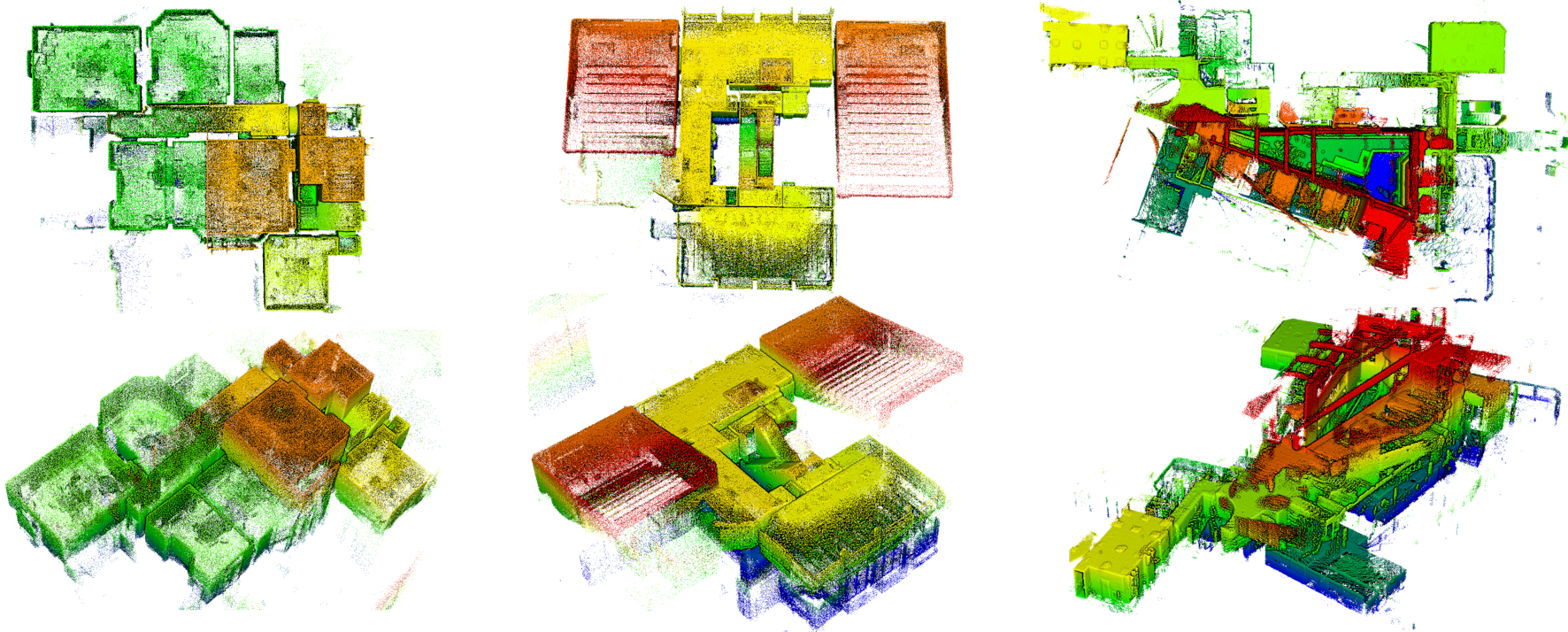}%
\caption{Our indoor datasets. The height direction is indicated by color: blue is the lowest level and red is the highest level.
        \textit{Left:} Small size George building. 
        \textit{Middle:} Medium size Thom building.
        \textit{Right:} Large size Information Engineering Building.
        }
\label{fig:dataset-envs}
\vspace{-1mm}
\end{figure*}

In this section, we describe experiments conducted on instance segmentation and descriptor networks. This is followed by real world experiments using InstaLoc as a complete localization system. Lastly, we demonstrate that the algorithm is robust to a changing number of prior map scans which indicates robust performance.

\subsection{Experimental Setup}
We use a fully labeled simulated dataset to train the instance segmentation and descriptor network. The dataset also holds 113 test scans for the instance segmentation network and 2123 test triplet instances for the descriptor network.

For the localization experiment, we collected an indoor environment dataset using a Ouster lidar OS-128 in small, medium, and large scale buildings. The dataset includes sequences in office rooms, meeting rooms, and social spaces as well as lecture theatres, staircases, and hallways.
Fig. \ref{fig:dataset-envs} shows the prior maps built with a lidar SLAM system. The SLAM poses are \SI{0.7}{\meter} apart so there are 147, 192, and 384 individual scans which form the final map for George, Thom, and IEB buildings. As an indication of size, the estimated map floor area for each building is around \SI{500}{\metre\squared}, \SI{1100}{\metre\squared}, and \SI{2000}{\metre\squared} respectively. However, in our localization experiments, the prior map is made up of a subset of registered scans that are spaced \SI{2.1}{\meter} apart. As the lidar sensor was running at \SI{10}{\hertz}, the localization system is triggered every ten scans - once per second. 

Tab. \ref{tab:results-recall-precision} presents specific details for each building. For example, the prior map of George Building consists of 32 scans, and the trajectory length is \SI{96}{\meter}. In total, 106 scans were queried. A detection is classified as being correct when the estimated pose is within \SI{0.2}{\meter} and the orientation is within \SI{10}{\degree} of the ground truth pose. Please note, there is no point cloud alignment step, such as Iterative Closest Point (ICP) refinement, and the pose estimation is from the instance correspondence matching.

\subsection{Results}
\subsubsection{Instance Segmentation Results}
\label{sec:inst-seg-results}
\begin{figure}
\centering
\includegraphics[width=0.60\columnwidth]{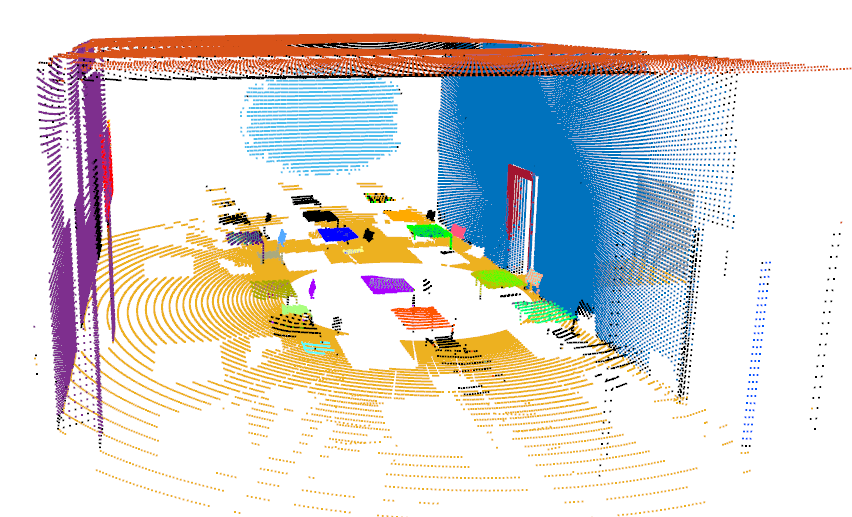}%
\includegraphics[width=0.40\columnwidth]{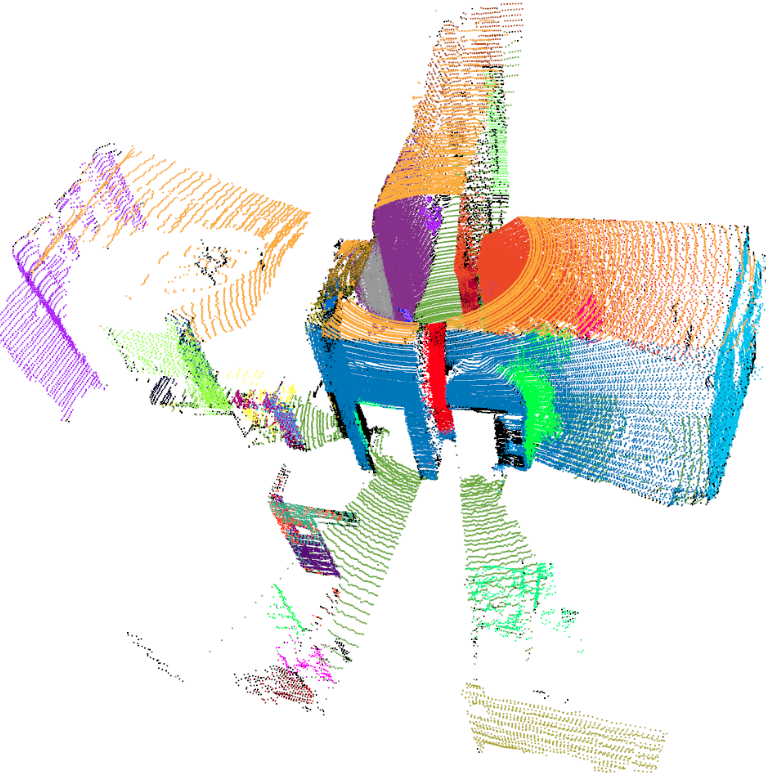}\\
\caption{Instance segmentation results. Left: Result with a simulated lidar scan. Right: Result with a real Ouster lidar scan. Random colors are assigned to each instance.}
\label{fig:inst-seg-result}
\vspace{-1mm}
\end{figure}

Fig. \ref{fig:inst-seg-result} shows two illustrations of instance segmentation results. The left side image of a scan is from the simulated dataset. In this classroom environment, each object instance has been assigned a random color. Chairs and tables are individually segmented and as well as each wall surface. Note that the door (colored in red) is partially segmented from the blue wall. In another example result, the right side image of a scan was captured in a hallway in George Building with the Ouster lidar. The hallway connects several rooms with lidar beams scanning into those rooms, which resulted in several partially scanned walls. The ceiling is accurately segmented (colored in orange), but the blue wall is mixed with one light green and one black segment. The imperfection in segmentation is expected as the current \sota instance segmentation method, SoftGroup, achieves an average precision (AP) of around \SI{54.5}{\percent} on indoor datasets such as the S3DIS dataset \cite{s3dis2016}. 

Unlike the S3DIS dataset, there is no visual color information for lidar points in our simulator synthesized data. In addition, our data contains a larger variety of spaces and objects than S3DIS, and some objects in the scans are scanned partially. Hence after applying the default SoftGroup on our synthesized data, it reaches \SI{39}{\percent} average precision across 13 classes. Our proposed improvement of incorporating lidar properties in \eqref{eq: adaptive-theshold} improved the Average AP from \SI{39}{\percent} to \SI{41}{\percent}, shown in Tab. \ref{tab:inst-seg-results}. Larger objects such as ceilings, floors, and walls have higher AP. Objects such as boards, windows, and doors, which are gathered under "other1" and "other2" in Tab. \ref{tab:inst-seg-results} have much low AP, less than \SI{20}{\percent}.

One key design consideration for our localization method to be able to deal with imperfect segmentation is to use all available instances with descriptors that can tolerate incomplete object point clouds.

\begin{table}
    \centering
    \begin{tabular}{ll|ll|ll}
    \toprule
         \textbf{Class} & \textbf{AP} & \textbf{Class} & \textbf{AP} & \textbf{Class} & \textbf{AP}\\
         \midrule
         ceiling & 0.923 & floor & 0.838 & wall & 0.565 \\
         column & 0.632 & beam & 0.367 & chair & 0.723 \\
         sofa & 0.402 & others1 & 0.144 & others2 & 0.163 \\
         \midrule
         \multicolumn{3}{c}{\textbf{Average AP}} & \multicolumn{3}{c}{41.3}\\
         \bottomrule
    \end{tabular}
    \caption{Average precision for each object class. others1 is the mean value of table, board, and window, others2 is the mean value of door, bookcase, and clutter.}
    \label{tab:inst-seg-results}
\end{table}

\subsubsection{Instance Descriptor Results}

\begin{figure}
\centering
\includegraphics[width=0.45\columnwidth]{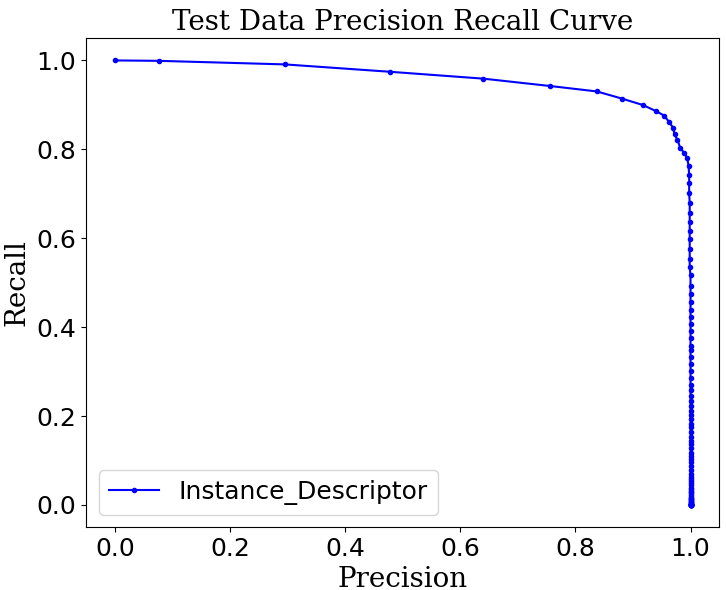}%
\hspace{1mm}%
\includegraphics[width=0.53\columnwidth]{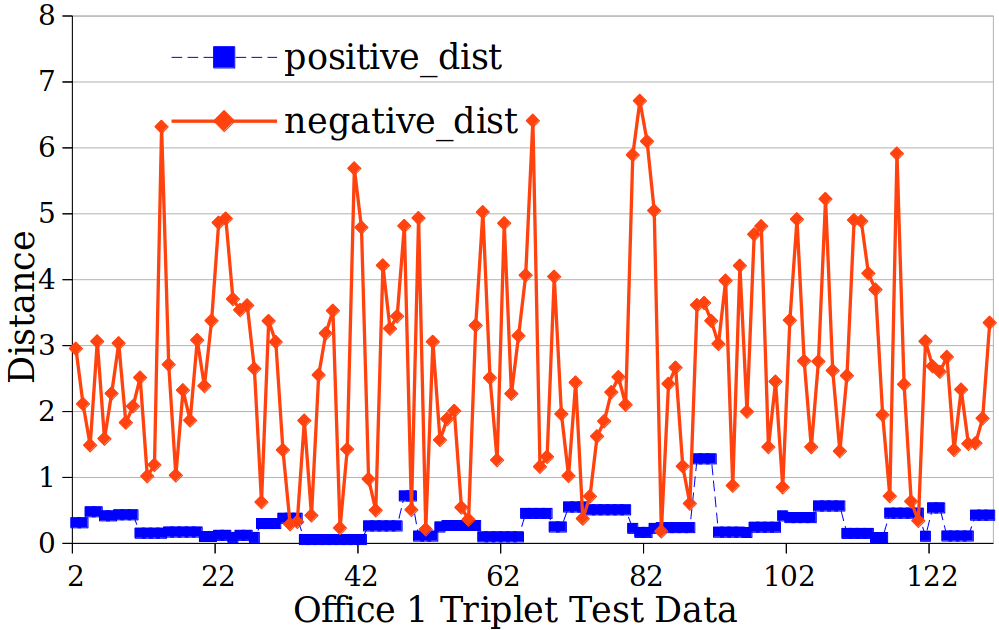}\\
\caption{\textit{Left:} A precision/recall curve for all test data in the descriptor network. \textit{Right:} A subset of the test data showing in blue the distance between the anchor and the positive instances. In red, the line shows the distance between the anchor and the negative instances.}
\label{fig:desc-pr}
\vspace{-1mm}
\end{figure}

In 3D point cloud learning, data representation and augmentation have a significant impact on achieving on best matching or labeling performance. We experimented with several approaches and found that centering individual instances and applying random rotations during data preparation can optimize learning results. We randomly eliminate \SI{20}{\percent} of the points in each instance and add random noise to lidar point positions during data preparation to improve descriptor robustness.

Fig. \ref{fig:desc-pr} (right) presents descriptor pairwise distances between the anchor, positive and negative instances in a subset of 120 test triplets. The blue lines correspond to smaller, positive distances (as desired). There is a clear separation between the typical positive and negative distances.

The graph in Fig. \ref{fig:desc-pr} (left) shows the precision and recall curve for the 2000 test triplets. At a descriptor distance ($\mathcal{L}_2$ norm) threshold of 0.56, the model can classify the instances with \SI{91.4}{\percent} precision and \SI{88.1}{\percent} recall. Here we purposely choose a smaller distance threshold to have higher precision as false positives are more detrimental to the localization system.
Our network is fast and efficient. For comparison, we tested on the Thom Building dataset. Averaged across all scans, ESM \cite{Tinchev2019esm} descriptor processed 30 segments in a scan in \SI{72}{\milli\second}, while InstaLoc descriptor network processed 30 instances in \SI{21}{\milli\second}.
In addition, our descriptor network can operate on any number of input points but ESM needs to downsample a segment to 256 points and SegMap uses fixed 3D voxel grid dimension of $32 \times 32 \times 16$ which compromises on detail.

\subsubsection{Localisation Results}

\begin{table*}
	\centering
	\vspace{2mm}
	\resizebox{\linewidth}{!}{%
		\begin{tabular}{c|c|cc|ccc|ccc|ccc}
			\toprule
			\textbf{Data} & 
                \textbf{Length} &
			\multicolumn{2}{c}{\textbf{Scan Number}} &
			\multicolumn{3}{|c}{\textbf{ESM* \cite{Tinchev2019esm}}} &
			\multicolumn{3}{|c|}{\textbf{SegMap* \cite{Dub2018SegMapRSS}}} &
			\multicolumn{3}{|c}{\textbf{InstaLoc (Ours)}} \\
			Building & (m) & Map & Query & Detect & Recall & Precision & Detect & Recall & Precision & Detect & Recall & Precision \\
			\midrule
                George & 96 & 32 & 106 & 12 & \SI{11}{\percent} & \SI{75}{\percent} & 28 & \SI{26}{\percent} & \SI{81}{\percent} & \textbf{56} & \textbf{\SI{49}{\percent}} & \textbf{\SI{93}{\percent}}\\
                Thom & 121 & 45 & 137 & 36 & \SI{26}{\percent} & \textbf{\SI{92}{\percent}} & 28 & \SI{30}{\percent} & \SI{83}{\percent} & \textbf{88} & \textbf{\SI{58}{\percent}} & \SI{91}{\percent}\\
                IEB & 253 & 98 & 211 & 29 & \SI{14}{\percent} & \SI{93}{\percent} & 27 & \SI{13}{\percent} & \SI{56}{\percent} & \textbf{94} & \textbf{\SI{42}{\percent}} & \textbf{\SI{95}{\percent}}\\
			\bottomrule
		\end{tabular}
	}
	\caption{Numerical summary table showing the performance of InstaLoc compared to two \sota benchmarks. The prior map is made of N scans and the query scan is the total number of scans queried. *: both methods have been adapted for better performance.}
        \label{tab:results-recall-precision}
\end{table*}

Fig. \ref{fig:teaser} shows a lidar scan that has been successfully localized within the ground floor of Thom building. Several object instances have been matched including walls, sofas, and flat planes (TV screens). The top section shows the matched instances within the query scan, and the bottom section shows the matched instance within the larger prior map. The estimated pose is indicated with a gold arrow. 

InstaLoc and two \sota baselines were tested with datasets from George, Thom, and Information Engineering Building (IEB). InstaLoc successfully detected 48 out of 106 scans in the prior map of George building, and all detections were correct according to the ground truth. Hence the recall rate is \SI{45}{\percent} and the precision is \SI{100}{\percent}. For Thom building, the recall rate was \SI{56}{\percent} but the precision was lower at \SI{86}{\percent}. The lower precision was largely due to the two near identical lecture theatre halls on the two sides of Thom building, shown in Fig. \ref{fig:dataset-envs}. This caused confusion in the localization system. Over the three sequences, the average recall was around \SI{47}{\percent}, and the average precision was about \SI{94}{\percent}. Note that all three datasets are for test, the segmentation and description networks have not seen them as they are trained on simulated lidar scans.

We selected two segment-based localization methods as comparative baselines, as they are most similar to our method. We modified the two algorithms to the best of our efforts to offer a fair comparison in indoor environments. 
In the Efficient Segmentation and Matching (ESM) paper\cite{Tinchev2019esm}, the authors used the Euclidean cluster extraction (ECE) method to segment the lidar scans. As it was originally designed for outdoor environments, objects in the scan are expected to be distinctly separated, especially after removing points corresponding to the ground. However, in an indoor environment, the ECE method cannot separate objects efficiently as walls and ceilings often become one segment. To mitigate this, we first calculate the curvature and remove high curvature points so there are distinct gaps between structured objects. After this, the ECE method can produce more reasonable segments.

The second algorithm we test is SegMap \cite{Dub2018SegMapRSS}. We first simplified the system by removing the lidar accumulating through the odometry system, as the lidar scans in our test are from a 128-beam lidar so it is very dense compared to 16 or 32-beam lidar used in their paper. More importantly, we used the incremental region growing method \cite{Dube2018IncrementalLoc} for segmentation, which computes local normals and curvatures for each point and uses these to extract flat or planar-like surfaces. After these two modifications, the system can operate in real time and have better segmentation performance.

However, even as we improved the segmentation method in both systems, there is still a limitation in their descriptor network. One factor is that it does not use sparse tensor networks, and as a result only a small and fixed number of points can be used as input.

A table presenting comparison results is shown in Tab. \ref{tab:results-recall-precision}, our approach outperforms the baseline methods by a factor of between two and four times in recall, and also achieved higher precision. In general, these systems tend to be tuned to prefer higher precision - for accurate and trustworthy localization.

We also considered ScanContext \cite{scancontext2018} for comparison, but its descriptor is too rudimentary to work in tight indoor spaces, as opposed to the road networks it was designed for.

\subsubsection{Varying the Size of the Prior Map}

\begin{table}
	\centering
	\vspace{2mm}
	\resizebox{\columnwidth}{!}{
		\begin{tabular}{c|cc|cc|cc}
			\toprule
                \textbf{Data} & 
                \multicolumn{2}{|c}{\textbf{Fewer Scans}} &
                \multicolumn{2}{|c}{\textbf{Default Density}} &
                \multicolumn{2}{|c}{\textbf{More Scans}} \\
			  Building & Map & R/P \% & Map & R/P \% & Map & R/P \% \\
			\midrule
                George & 22 & 30 / 94 & 32 & 45 / 100  & 48 & 49 / 84 \\
                Thom & 33 & 45 / 97 & 45 & 56 / 86  & 60 & 54 / 86 \\
                IEB & 68 & 30 / 100 & 98 & 41 / 97  & 125 & 42 / 93 \\
			\bottomrule
		\end{tabular}
	}
	\caption{Ablation study: varying the number of scans used for the prior map. The same number of query scans are used as Tab. \ref{tab:results-recall-precision}. R and P are the recall and precision values respectively.}
        \label{tab:results-vary-priop-map}
\end{table}

As a robustness test, we conducted experiments to see how the number of individual scans in a prior map can affect localization results. Intuitively, as the number of prior map scans is reduced, the localization detection rate should reduce. Shown in Tab. \ref{tab:results-vary-priop-map}, the middle column is the baseline which has the same configuration as \ref{tab:results-recall-precision}, and the columns left and right of it have either an increased or decreased number of prior map scans. With the decreased number of prior map scans, there is a slight reduction in recall values, but no negative effect on precision. This demonstrates the robustness of our system to changes in the number of prior map scans.

\subsection{Limitations}
\begin{figure}
    \centering
    \includegraphics[width=\columnwidth]{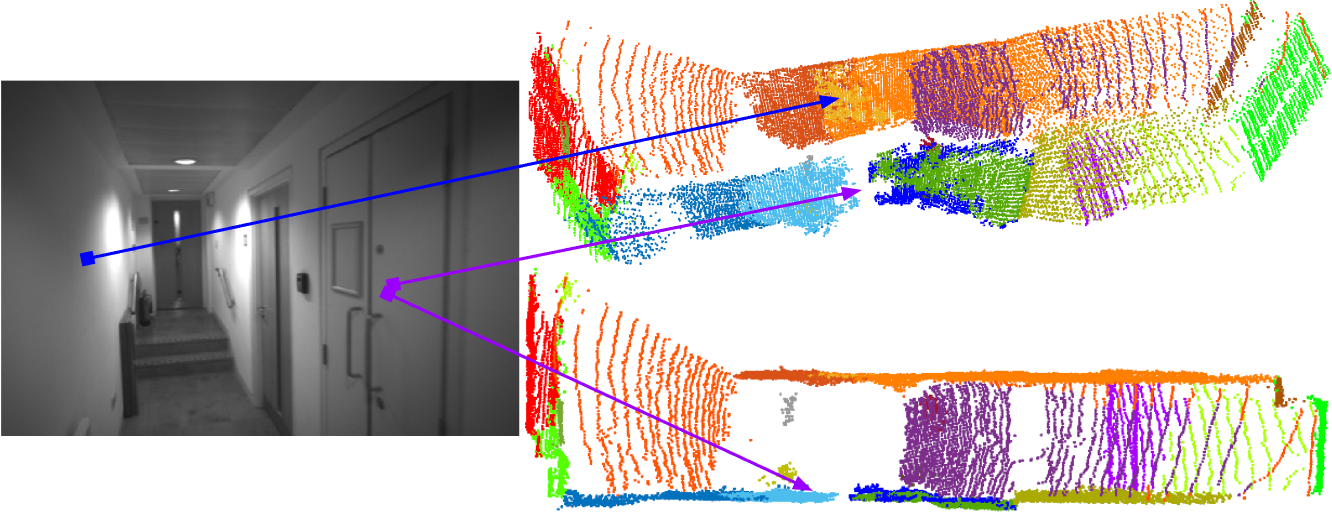} \\
    \caption{An example of inferior instance segmentation within a corridor from two different viewpoints. In the side view scan, both the left and right walls are being over-segmented. In the top view scan, the points near the sensor origin ($<$\SI{1.0}{m}) have much higher noise, resulting in uneven wall surfaces.}
    \label{fig:diff-corridor}
\end{figure}

As mentioned in Sec. \ref{sec:inst-seg-results}, the precision of the instance segmentation can directly impact the performance of the localization system. Since the descriptor network has already reached high precision and recall values, good instance segmentation is a key way of improving overall localization recall performance. In our experiments, the instance segmentation network performs well in structured and enclosed spaces such as theatres, classrooms, offices, etc. However, it performs much more poorly in corridors and staircases, especially when there are embedded small objects inside the walls, such as handrails.

As shown in the camera image in Fig. \ref{fig:diff-corridor} (left), there were fire extinguishers, radiators, and cupboards along the corridor walls. We did specifically use examples of hallways and corridors in our training dataset. While that did improve performance, the results still have room for improvement. This might be due to the inconsistent point cloud density on the walls and the noisier lidar measurements from the Ouster lidar at close distances. An example of this issue is shown in Fig. \ref{fig:diff-corridor} (right) in a corridor. This is a topic for future work.

\section{Conclusion}
\label{sec:conclusion}

In this paper, we proposed a fast and accurate lidar localization approach. InstaLoc learns to segment and describe different object instances in a scene. Two networks inside are joined together to recognize and then describe the individual objects. InstaLoc can localize with between two to four times as many matches as two \sota baseline methods while retaining high levels of precision. In future work, we want to improve the localization performance in hallways and corridor spaces. Moreover, we intend to combine visual information with lidar measurements in instance segmentation. As importantly, we aim to extend InstaLoc to outdoor environments and also to be independent of the type of operating environment. Lastly, we will add flexibility to InstaLoc to work with sparse lidar scans from different lidars.

\section*{Acknowledgments}
This research was partly funded by the Horizon Europe project DIGIFOREST (Grant ID 101070405), UKRI/EPSRC ORCA Robotics Hub (EP/R026173/1), and a Royal Society University Research Fellowship (Fallon).

\balance
\bibliographystyle{plainnat}
\bibliography{references}



\section*{Supplementary material}

# Appendix: Further Implementation Details

This appendix discusses further implementation details of the InstaLoc method and acts as an extension to Sec. IV of the main paper. In particular, we will discuss the training procedures, data generation and augmentation, and our triplet mining strategy.

## 1 Panoptic Segmentation Network

For panoptic segmentation of the lidar scan, we adopted SoftGroup method of Vu, Thang et al[1], a state-of-the-art method for 3D point cloud instance segmentation. SoftGroup offers a model pertained on the S3DIS dataset. Each scan in the S3DIS dataset is a high accuracy static 3D scan of a room with millions of points. Hence we need to adapt the network and model to better suit the sparser point clouds from 3D automotive lidars such as Ouster or Velodyne. Details of the algorithmic adaptation are in Sec III B. The radius threshold $\rho_i$ in Equa. (2) is added into CUDA processing for each point.

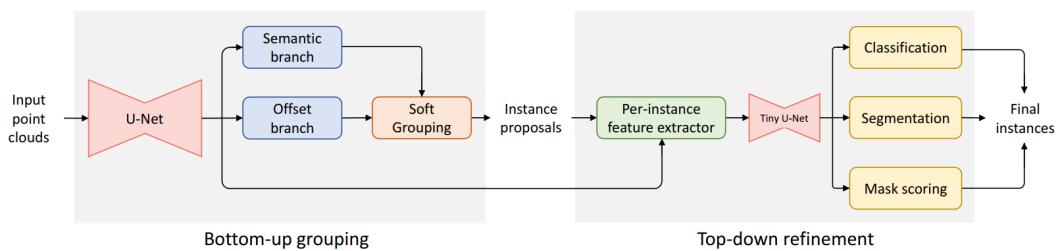

Figure 1: SoftGroup panoptic segmentation architecture

### 1.1 Generating Simulated Scans

Extending the description presented in Sec. IV A, we purchased the 3D model assets [2] from the Unreal Engine marketplace and built a six-room building space. Within this space we simulated the lidar scans shown in Fig. 4. We enabled complex collision properties in the Unreal Engine to obtain object details such as object instance and class for each simulated lidar points - not just just their bounding boxes. Similarly to the S3DIS dataset, we assign the objects to 13 classes. Tab. 1 shows the number of objects in each class.

- Movable: board, bookcase, chair, sofa, table
- Fixed: beam, ceiling, column, door, floor, wall, window

Note that objects are classed as clutter, which could be both movable or fixed, but are typically movable.

| Level | beam | board | bookcase | ceiling | chair | clutter | column | door | floor | sofa | table | wall | window | Total |
|---|---|---|---|---|---|---|---|---|---|---|---|---|---|---|
| Level 1 | 4 | 4 | 10 | 1 | 34 | 119 | 3 | 8 | 4 | 0 | 37 | 18 | 9 | 251 |
| Level 2 | 4 | 10 | 11 | 1 | 39 | 113 | 4 | 8 | 4 | 1 | 32 | 20 | 7 | 254 |
| Level 3 | 4 | 12 | 7 | 1 | 35 | 109 | 4 | 8 | 6 | 4 | 33 | 30 | 7 | 253 |
| Level 4 | 1 | 16 | 8 | 1 | 35 | 109 | 3 | 11 | 6 | 6 | 26 | 20 | 9 | 251 |
| Level 5 | 5 | 14 | 3 | 1 | 59 | 65 | 6 | 9 | 5 | 2 | 50 | 16 | 13 | 248 |
| Level 6 | 7 | 5 | 4 | 1 | 41 | 98 | 8 | 7 | 11 | 5 | 38 | 16 | 13 | 254 |

Table 1: Number of objects in each class in the simulated environments

### 1.2 Data Augmentation

Similarly to many point cloud segmentation methods, we augmented the scans by adding Gaussian noise of 1cm along each of the three dimensions, randomly mirror flipping each scan, and applying random yaw rotations.

---

[1] Vu, Thang et al. "SoftGroup for 3D Instance Segmentation on Point Clouds." 2022 IEEE/CVF Conference on Computer Vision and Pattern Recognition (CVPR) (2022): 2698-2707.

[2] https://www.cgtrader.com/3d-models/interior/interior-office/low-poly-interior-13-office

## 1.3 Training

We captured 270 simulated scans for training, validation, and testing, and split them into 90 scans each. The training step aims to fine-tune the pre-trained S3DIS model to adapt to the sparser lidar scans. The original S3DIS dataset contains 271 scans so we use a smaller number of simulated scans for fine-tuning training.

To fine-tune the pre-trained model, we adapted a two-step training process. First, we froze the modules in the top-down refinement section (highlighted in color) and only trained the U-Net, semantic branch, and offset branch modules. Second, we froze the aforementioned modules and fine-tuned the modules in the tiny U-Net, classification, segmentation, and masking scoring. In this way, we can maintain all the weights in the pre-trained model and adjust them for sparse 3D lidar scans.

# 2 Instance Descriptor Network

## 2.1 Generate Triplet Data

Extending the description in Sec IV A, we obtain instance triplets from the simulated lidar scans in 1.1. So as to achieve view invariance in the instance descriptor, we generate pairs of Lidar scans from nearby poses in each room, 2 m apart with a $10°$ rotation around the yaw axis. The algorithm to generate triplets is described below. In total, we generated 9900 triplets to train the descriptor network.

---

**Algorithm 1:** Generate triplets for a pair of lidar scans

**Result:** N number of triplets
Segment objects in each scan by using their ground truth object instance ID;
Bin the objects based on their semantic label;
**while** *each semantic label* **do**
    `anchor`: Pick an object from scan 1;
    `positive`: Pick the same object from scan 2, using the instance ID;
    `negative`: Randomly pick 4 other objects: from the same class in scan 1 and 2; from a different class in scan 1 and 2;
    **return** 4 x (anchor, positive, negative)
**end**

---

## 2.2 Data Augmentation

For each instance in the simulated triplet data, we follow the same procedure as Appendix 1.2. We noticed the simulated lidar scans have very even spacing between points and little noise in the scan line pattern compared to real lidar scans, so we randomly eliminate $20\%$ of the points in each instance to mimic such effect.

## 2.3 Training

The train, validate and test split for the triplet network was chosen to be around 60, 20, and $20\%$. As the size of each triplet is relatively small compared to the original point cloud, we train and test 8 triplets in a batch on our mobile GPU. Note that during inference we combine all instances from each point cloud as one batch for the forward pass.

# 3 Sim to Real Transfer

In general, the sim-to-real domain gap between lidar scans is smaller than with images. Images depend on lighting, reflection, camera angle, and object texture making transfer much more challenging. By contrast, 3D lidar scans only contain geometric information (XYZ point coordinates) which is much more amenable to direct transfer and training with a smaller amount of samples. In addition, we applied data augmentation in Appendix 1.2 and 2.2 to narrow the gap between the simulated point clouds and the real point clouds.



\end{document}